\documentclass{article}
\usepackage{spconf,amsmath,graphicx}
\usepackage{multirow}
\usepackage{bbding}
\def\@fnsymbol#1{\ensuremath{\ifcase#1\or \dagger\or* \or \ddagger\or
		\mathsection\or \mathparagraph\or \|\or **\or \dagger\dagger
		\or \ddagger\ddagger \else\@ctrerr\fi}}
\title{Intra-clip Aggregation for Video Person Re-identification}
\name{Takashi Isobe$^\dagger$~~~~~~~Jian Han$^\dagger$~~~~~~~Fang Zhu$^\ddagger$~~~~~~~Yali Li$^{\dagger}$~~~~~~~Shengjin Wang$^{\dagger}$\textsuperscript{\Envelope}}
\address{$^\dagger$Beijing National Research Center for Information Science and Technology,\\ Department of Electronic Engineering, Tsinghua University, Beijing 100084, China\\$^\ddagger$MetroTech Center, Brooklyn, New York University, 11201, USA}

\begin{document}
%\ninept
%
\maketitle
\begin{abstract}
Video-based person re-identification has drawn massive attention in recent years due to its extensive applications in video surveillance. While deep learning-based methods have led to significant progress, these methods are limited by ineffectively using complementary information, which is blamed on necessary data augmentation in the training process. Data augmentation has been widely used to mitigate the over-fitting trap and improve the ability of network representation. However, the previous methods adopt image-based data augmentation scheme to individually process the input frames, which corrupts the complementary information between consecutive frames and causes performance degradation. Extensive experiments on three benchmark datasets demonstrate that our framework outperforms the most recent state-of-the-art methods. We also perform cross-dataset validation to prove the generality of our method.

\end{abstract}

% poor representation of discriminative clip-level descriptors # In training use learning clip-level embedding. In testing use represention of discriminative clip-level descriptors


%the learned clip-level embedding improves the state-of-the-art performance on two public video re-ID benchmarks. To confirm the advantage of synchronized transformation, we conduct ablation study with different synchronized transformation operations. Further, we show the promising result of the proposed method on action recognition task.

%In this paper, we propose a simple yet effective Synchronized data augmentation scheme for video person re-ID. This scheme also shows its effectiveness on other temporal aggregation task (~\emph{eg.} action recognition). Inconsistent intra-clip augmentation  To tackle the above-motioned problems, we design a novel framework for video-based person re-id, which consists of two main modules: Synchronized Transformation (ST) and Intra-clip Aggregation (ICA). The former module augments intra-clip frames with the same probability and the same operation, while the latter leverages two-level intra-clip encoding to generate more discriminative clip-level features. To confirm the advantage of synchronized transformation, we conduct ablation study with different synchronized transformation scheme.  Extensive experiments on three benchmark datasets demonstrate that our framework outperforming the most of recent state-of-the-art methods. 

%
\section{Introduction}
\label{sec:intro}

Person re-identification (re-ID) aims to recognize the same identity in different images or videos captured by different cameras distributed at separated physical locations. 
% With the emergence of deep learning methods within recent years and their influence on the computer domain, video person re-ID is in high demand for video surveillance and has driven significant progress. Specifically, video person re-ID is a generalized image person re-id problem. Given a interest tracklet (query) of one person and a gallery set which contains a number of candidate tracklets observed by another cameras, the system aims to calculate similarity metric between two clip-level descriptors and find the query person in gallery set. 
 In contrast with image person re-ID~\cite{sun2018beyond,sun2017svdnet,li2018harmonious,isobe2021towards}, video person re-ID is more robust to noise. Both spatial information across positions and temporal information across frames can be used to represent clip-level features. The previous works~\cite{liao2018video,liu2017quality} exploit motion estimation either implicitly (\textit{e.g.} gait) or explicitly (\textit{e.g.} optical flow) to represent a video sequence. But those works are not optimal for video person re-ID. Inaccurate motion estimation, especially when there is occlusion or parallax, deteriorating the final performance. Besides, those methods often suffer a heavy computational load. 
%  ~\cite{zhou2017see,liu2018video,song2018region}
 \begin{figure}[t]
\includegraphics[width=0.48\textwidth] {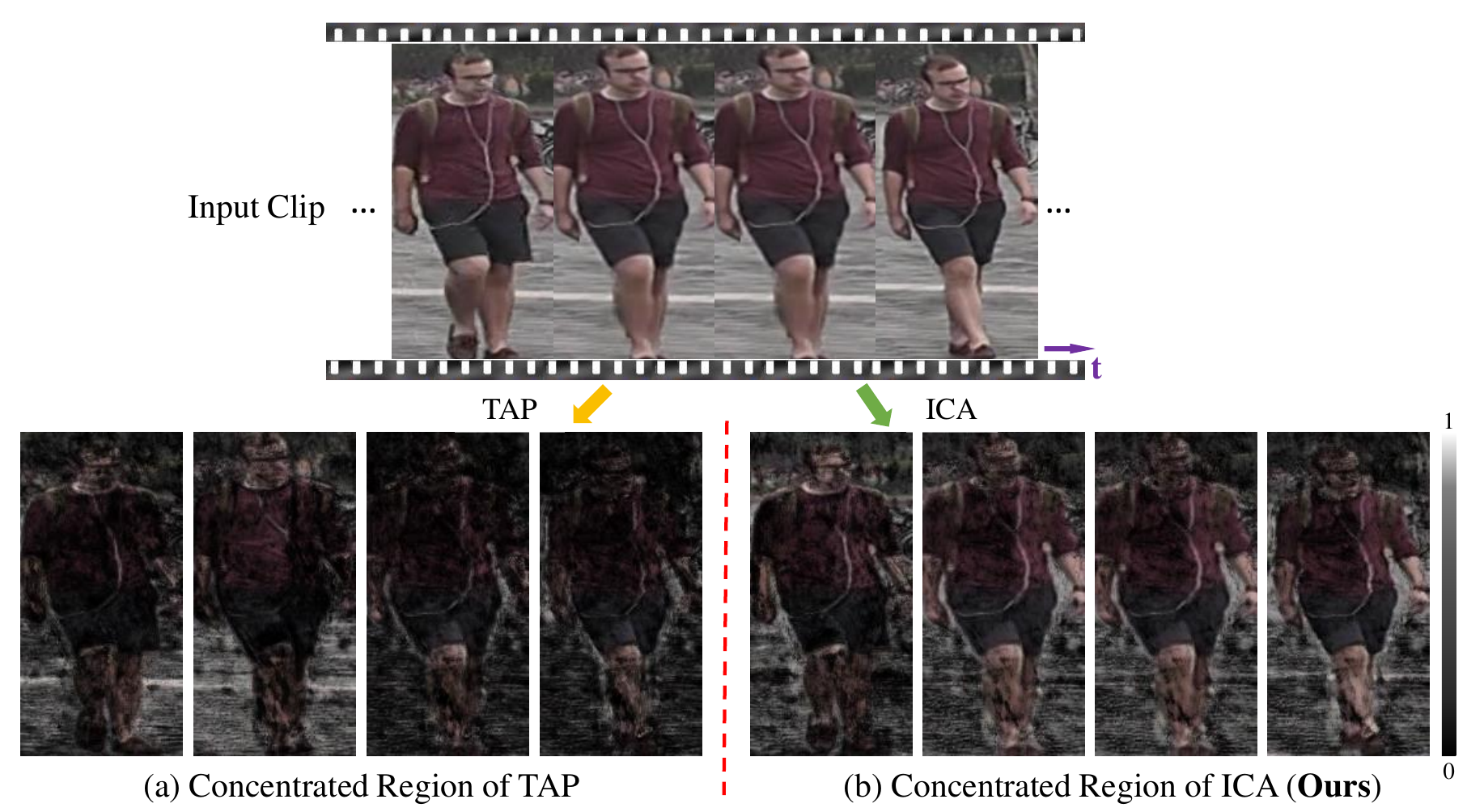}
\caption{Visualization of the concentrated regions for the intra-clip frames. For fair comparison, we use same feature extraction backbone, \textit{i.e,} ResNet50~\cite{he2016deep}, but with different temporal aggregation strategy. (a) and (b) adopt Temporal Average Pooling (TAP) and the proposed hierarchical aggregation strategy, respectively. In (a), the activated maps have a scattered distribution with less meaningfulness. In (b), the activated maps are more concentrated and meaningful around body parts. The activated maps are obtained by Grad-cam~\cite{selvaraju2017grad}.}
\label{fig1}
\end{figure}
To represent a clip-level descriptor while maintaining a low computational cost, temporal pooling has been widely used in recent works~\cite{gao2018revisiting,li2018diversity}. They perform generic or weighted average pooling in the end of the network to aggregate intra-clip features across time. However, temporal pooling is a linear operation which is limited to capture the specific features of a video sequence. In this paper, we propose a novel Intra-Clip Aggregation (ICA) module in order to effectively integrate the clip-level features. Specifically, ICA is a cascade structure which consists of a learnable block followed by a temporal average pooling layer. The critical innovation of the ICA is hierarchically aggregating the intra-clip features with both linear and non-linear operations. The linear operation is used to generate global features, and then we apply a non-linear block to describe the most important semantic concepts of clip-level features. 
% Fig.1 shows the limitation of the previous aggregation method. (a) involve noise to the clip-level features. (b) mitigate the informative information of other frames. (c) is our method, which leverage the interaction information of other frames to compensate the lost information.

% full integrate the clip-level information
% from the temporally pooled feature vectors.
% It can not change the intrinsic property of the aggregated 
% which aggregate the frame-level features and then use a group of convolution kernels to represent more discriminative clip-level features. Instead of o ICA module , which incorporate temporal pooling and 
% Although these approaches with temporal pooling shows promising performance, but the learned clip-level embedding is weaken due to In this paper, we proposed a discriminative intra-clip frame embedding block which map the aggregated features 
%  Moreover,which . Thus the aggregated clip-level features may not be sufficiently discriminative for identifying different persons. 
%  make the CNN better leveraging the interaction information among frames in stead of only describe the given dataset.
% 1. introduce of data augmenataion
% 2. temporal cues 
% 3. lost information
% 4. overffit to noise
% 5. details
% 6. 1) kept unchanged; 2) synchronized augmented.

Data augmentation, an explicit form of regularization, has been widely used in the training process of deep neural network.
% To reduce the risk of overfitting, data augmentation is a effective way by randomly transferring or noising the original images. 
The general data augmentation approaches such as random cropping, flipping as well as erasing~\cite{krizhevsky2012imagenet,zhong2017random} work well on image person re-ID task by randomly transferring or noising the original images. The previous works~\cite{gao2018revisiting,su2018spatial,liu2017quality} treat video person re-ID as a generalized image person re-ID task. They apply data augmentation operation asynchronously on the input frames. In such process, each frame is transformed with a random probability, which introduces excessive noise and consequently corrupts the temporal cues.
% Asynchronous transferring or noising a video sequence may introduce extremely additional noise and corrupt the temporal cues.may lost too much spatio-temporal information for a tracklet
For example, randomly flipping each frame will result in misalignment. In addition, randomly erasing a region of pixels of each frame will cause too much spatial-temporal information loss. In this case, the model may be confused to fully utilize the interactive information to represent informative clip-level features and perform poorly in the presence of real-world noise, \textit{e.g.}, occlusion, lighting and motion blur. In this paper, we propose a video-based data augmentation approach for video person re-ID, which is a temporal extension of commonly used image-level data augmentation techniques. The proposed video-based data augmentation is easy to implement and meanwhile yields consistent improvement over three challenging video person re-ID benchmarks. 
% and shows promising performance on cross-data validation and action recognition task.

To sum up, our contributions are listed as following:
\begin{itemize}
\item
We revisit data augmentation for video person re-ID task, and propose a novel video-based data augmentation strategy to strengthen the representation ability and the generality of the learned model. It can be adopted on various existed image-based data augmentation approaches.
% Our method can effectively retain the complementary information between consecutive frames and randomly change the noise which exist in underlying data distribution making the learned model robust to real-world noise.
\item 
% We introduce a novel cascaded module, \textit{i.e.}, the ICA module, to aggregate the intra-clip features hierarchically with linear and non-linear mapping. The non-linear mapping block can improve the representation capability of CNN based temporal pooling methods, also archives impressive result on cross-data validation showing the generality of the learned model. 
We propose a novel cascade temporal integration pipeline which effectively integrates the intra-clip features in a hierarchical manner. 
% \item
% We extent the commonly used image-level data augmentation to augment the frames in one clip synchronously. The proposed Synchronized Transformation  can effectively retain the complementary information between consecutive frames and randomly change the noise which exist in underlying data distribution making the learned model robust to real-world noise.
% \item 
% Extensive experiments show the effectiveness of the proposed temporal modeling framework. By unifying the synchronized augmentation and ICA, our method outperforms recent proposed state-of-the-art methods on two challenging benchmark datasets, ILIDS-VID~\cite{wang2014person} and MARS~\cite{zheng2016mars} without re-ranking. To better understand the generality of the learned model, We also perform cross-dataset validation on PRID2011~\cite{hirzer2011person}. Further, we show that such video augmentation strategy can also yield consistent improvement in other intra-clip aggregation task such as action recognition, which can be regarded as a spatio-temporal integration problem.
\item
Our model outperforms the state-of-the-art methods on ILIDS-VID~\cite{wang2014person} and MARS~\cite{zheng2016mars} benchmarks by a large margin. The impressive result on cross-data validation shows the generality of the proposed method.
% Further, we show that such video augmentation strategy can also yield consistent improvement in other intra-clip aggregation task such as action recognition, which can be regarded as a spatio-temporal integration problem.
\end{itemize}

\section{Methodology}
% \subsection{Approach Overview}

\begin{figure}[t]
\includegraphics[width=0.48\textwidth] {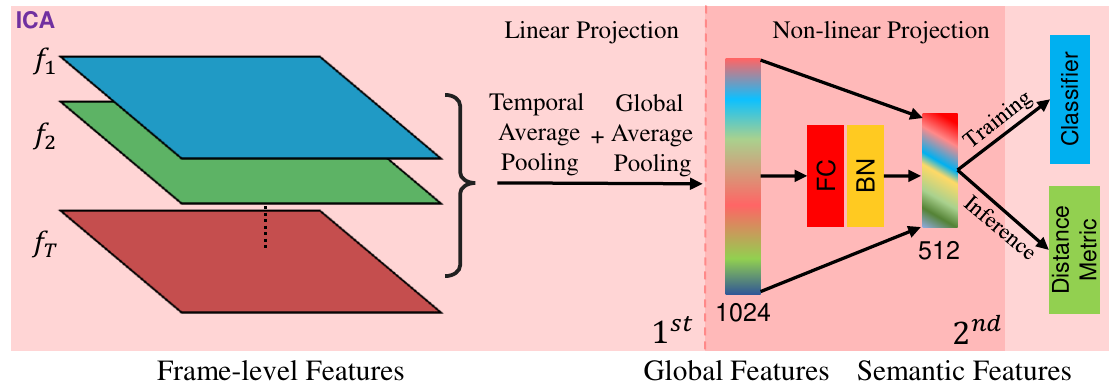}
\caption{The proposed ICA module which integrates the intra-clip information with a hierarchical pattern.}
\label{ica}
\end{figure}

% The overall framework is illustrated in Fig.2.
In training process, the intra-clip frames within a mini-batch either are augmented synchronously or remain unchanged, and then a base network is applied to extract the frame-level features. Finally, ICA module deeply integrates intra-clip features in a hierarchical manner to represent the clip-level pedestrian descriptors. Fig.~\ref{ica} shows the proposed ICA module. The main contributions of the overall framework are two parts: 1) video-based data augmentation; 2) hierarchically temporal integration module. More details about synchronous data augmentation and ICA module are presented in Sec.2.1 and Sec.2.2, respectively.

% Temporal pooling is an simple yet effective approach for temporal aggregation task. However, there are two drawbacks 1) linear mapping not change the property of clip-level features. 2) In the presence of occlusion, temporal pooling suffers sever performance degeneration.  

% the composite features derived from temporal pooling lack of discriminative representativeness due to the fact that the temporal pooling cannot learn from clip-level feature. To handle this problem, we proposed an Intra-clip Aggregation (ICA) module that aggregates the frame-level features and then uses a group of convolution kernels to
% learn the distinctive clip-level features. In particular, compared with simple temporal pooling baseline, our method generates more discriminative clip-level features while maintains fewer computational load. We will introduce  The other contribution of our method is clip-level data augmentation techniques. Video frame appearance plays an important factor in temporal pooling methods. However, frame-based data augmentation techniques randomly augmented each frame that make the frame unaligned and bring in additional noise factor for each frame. To address this problem, we come up with clip-level initialization techniques, which synchronously augment intra-clip frames. Overall, our person re-id matching model consists of two iterative procedure: (1) syncronized data augmentation and (2) generalized clip-level feature aggregation, as elaborated in the following.

\subsection{Synchronous Data Augmentation}
In this subsection, we improve the data augmentation strategy for video person re-ID task. Commonly used image-level data augmentation~\cite{krizhevsky2012imagenet,zhong2017random} approaches perform well on image person re-ID by suppressing the underlying noise such as camera intrinsic noise and background noise. However, as for a video sequence, asynchronous data augmentation may introduce unnecessary noise corrupting the temporal cues of intra-clip frames, which leads to the result that the model poorly resists the noise from the real world and may be confused about how to utilize the intra-clip complementary information. 
% Moreover, the intra-clip complementary information has been weaken as superposed involve more asynchronous data augmentation approaches.  
%  \begin{figure}[h]
% \includegraphics[width=0.5\textwidth] {pic/erasing.pdf}
% \caption{Example of asynchronous random erasing and synchronized random erasing.}
% \label{syn}
% \end{figure}
% For video person re-ID, the CNN is expected to learn how to use the potentially complementary information among consecutive frames and generate robust features of pedestrian tracklet. 
To address aforementioned drawbacks, we propose a novel video-based data augmentation strategy, termed as Synchronous Data Augmentation (SDA). Our method can effectively preserve the complementary information among consecutive frames and change the underlying noise among frames synchronously, which helps the network to learn a discriminative distance metric and better utilize the interactive information among frames. In training, the intra-clip frames within a mini-batch randomly undergo either of the two operations: 1) remaining unchanged; 2) being synchronously transformed with commonly used data augmentation techniques such as random flipping~\cite{krizhevsky2012imagenet} and random erasing~\cite{zhong2017random}. We formulate the operation of asynchronous transformation and the proposed asynchronous transformation as following. For simplicity, we formulate one data augmentation process as example. \\ 
%  which exist in underlying data distribution,
% In this section, we describe the temporal data augmentation (temporal Transform) techniques in details. Our scheme expands the image-level augmentation techniques to clip-level, which uniformly transforms all frame of a given sequence, just like artificially cropping or erasing the same patch of each frame.
asynchronous transformation:
% \begin{equation}\label{init_at}g
%     \{\psi_k(\cdot)\}_{k=1}^T\sim Rand(0,1)
% \end{equation}
\begin{equation}\label{sa}
{T_{at}}\{f_1,f_2,...,f_T\} = \{\psi_k(f_k)\}_{k=1}^T
% {T_{at}}\{f_1,f_2,...,f_T\}=\left\{
% \begin{array}{rcl}
%   \{f_1,f_2,...,f_T\}& & P\\
%      \{\psi_k(f_k)\}_{k=1}^T  &  &1-P
% \end{array}\right
% % {\rm T_{at}}\{f_c^1,f_c^2,...,f_c^T\}=\{\Psi(\Psi_2\mid\Psi_1(f_c^1)),\Psi_3(\Psi_2\mid\Psi_1(f_c^2)),...,\Psi_3(\Psi_2\mid\Psi_1(f_c^T))\}
\end{equation}
% \begin{equation}
%     {T_{at}}\{f_1,f_2,...,f_T\}=
%     \begin{cases}
%     \{f_1,f_2,...,f_T\}& P,\\
%     \{\psi_k(f_k)\}_{k=1}^T &1-P.
%     \end{cases}
% \end{equation}
synchronous transformation:
% \begin{equation}\label{init_st}
%     \psi(\cdot)\sim Rand(0,1)
% \end{equation}
\begin{equation}\label{st}
% % \begin{equation}
%     {T_{at}}\{f_1,f_2,...,f_T\}=
%     \begin{cases}
%     \{f_1,f_2,...,f_T\}& P,\\
%     \{\psi(f_k)\}_{k=1}^T & 1-P.
%     \end{cases}
% \end{equation}
{ T_{st}}\{f_1,f_2,...,f_T\} = \{\psi(f_k)\}_{k=1}^T
% {\rm T_{at}}\{f_c^1,f_c^2,...,f_c^T\}=\{\Psi(\Psi_2\mid\Psi_1(f_c^1)),\Psi_3(\Psi_2\mid\Psi_1(f_c^2)),...,\Psi_3(\Psi_2\mid\Psi_1(f_c^T))\}
\end{equation}
where $\Psi(\cdot)$ denotes the operator of data augmentation. For asynchronous data augmentation ${T_{at}\{\cdot\}}$, the operation $\{\Psi_k(\cdot)\}_{k=1}^T$ is randomly changed over the input $T$ frames. For synchronous data augmentation ${T_{st}\{\cdot\}}$, all frames are applied with the same operation $\Psi(\cdot)$. $\{f_k\}_{k=1}^T$ denotes $T$ frames of a tracklet.

% As illuminated in Figure 2, the proposed temporal augmentation techniques are the first part of overall framework. 
In this paper, we incorporate three types of augmentation approaches, \textit{i.e.}, random cropping, flipping and erasing. The transformation probability of each operation is fixed along temporal axis, \textit{i.e.}, cropping size, rotating angle and erasing region.
% we conduct exploratory experiment on Sec.4.2 and present a group of visualization result to support our analysis.
% \begin{figure}[t]
% \begin{center}
% \includegraphics[width=0.9\textwidth] {images/ICA.png}
% \caption{Illustration of ICA module. The left part shows two-type of pooling operation (temporal pooling and global average pooling) to generate a temporary clip-level feature. The right part is a learnable block, which are used to generate more high-level intra-clip features. We designed the learnable block as bottleneck structure.}
% \end{center}
% \end{figure}
% In Fig~\ref{syn}, (a) and (b) illustrate the frames of one clip are processed by asynchronous and synchronous augmentation, respectively. 
\subsection{Intra-clip Aggregation Module}
The key idea of ICA module is to capture the important semantic concepts of clip-level features. Based on such motivation, ICA is designed as a cascade structure, which can fully integrate clip-level information in a hierarchical manner. 
ICA takes the frame-level features as input, and performs average pooling to generate clip-level global features in preliminary fusion, which can be expressed as: 
% \begin{equation}\label{label}
%     Y_{1,c,h,w}=F_{tp}(X_{t,c,w,h})=\frac{1}{T}\sum_{t = 1} ^T X_{t,c,w,h}
% \end{equation}
\begin{equation}\label{label}
   Z_{1,c,1,1}=\frac{1}{WHT}\sum_{w = 1} ^W \sum_{h = 1} ^H\sum_{t = 1} ^T X_{t,c,w,h}
\end{equation}
Where $ X_{t,c,w,h} $ is temporally concatenated frame-level features. $Z_{1,c,1,1}$ is clip-level global features obtained by linear projection.
% It is notable that intra-clip pooling does not change the dimension of frame channel, which means this operation not generated any high-level features. 
Subsequently, the above $Z_{1,c,1,1}$ is further integrated with a high dimensional feature projection block, which can be formulated as:
% Thus, we concatenate a learnable module after that to encode more distinctive clip-level features. The formulation of clip-level encoding as following:
\begin{equation}\label{label}
   Y_{\tilde{c}}=G_{c\to \tilde{c}}(Z_{1,c,1,1})
\end{equation}
Where $Y_{\tilde{c}}$ stands for the clip-level semantic embedding. $G_{c\to \tilde{c}}(\cdot)$ denotes the non-linear projection. To reduce the computational cost, we design $G_{c\to \tilde{c}}(\cdot)$ as a bottleneck structure, which is composed of fully-connected (FC) layer and batch normalization (BN)~\cite{ioffe2015batch}. More details about the parameter setting and structure of ICA can be found in Fig.~\ref{ica}.
% is composed of $1\times1 convolutional layer$ with 256 channles followed by a Batch Normalization (BN)~[cite] and Leaky Relu~[cite].     .  

In comparison with the most existing methods~\cite{gao2018revisiting,li2018diversity}, which typically adopt a temporal pooling layer in the end of the network to represent the clip-level features, our method is able to generate more discriminative clip-level features by leveraging the interactional information among consecutive frames (see Fig.~\ref{fig1}). The impressive experimental results about synchronous data augmentation and ICA module are presented in Sec.3.
%  frames (see Fig.~\ref{fig1})
%  The of ICA is deeply extract the underlying clip-level features  .
% Our method is to extract the interactional information among consecutive frames in an integration mode and then deeply learn the underlying clip-level features to generate more discriminative clip embedding .
% The non-linear projection block can improve the representation capability of CNN based temporal pooling methods,
% The output of the first linear aggregation is further integrated by a learnable block, which performs a non-linear projection to represent a semantic clip-level features. The 
% Such non-parametric learning can not learn the underlying clip-level distribution.
% The global features are further aggregated by a non
% Specifically, ICA takes frame-level features as input, and then perform linear encoding and non-linear projection orderly to represent the clip-level descriptors. 

% In this section, we describe the structure details of the proposed Intra-clip Aggregation (ICA) module. Figure 2 shows that our ICA module consists of two sub-modules. The left part exploit two sequential non-parameter encoding operation to generate a temporary clip-level features, and another part utilize a clip-level encoding operation to reconstruct the clip-level feature into high-class representation feature. We formulate the intra-clip pooling operation as following:

%-------------------------------------------------------------------------
\section{Experiments}
% In this section, we comprehensively explore the proposed clip-level data augmentation technique and temporal aggregation module on popular benchmark video re-id datasets.

% % Firstly, we introduce the experimental setting and implementation details in Sec 4.1, and then conduct ablation study and present a group of visualization to support the ablation analysis in Sec 2.2. In Sec 4.3, we compare the performance of our method with state-of-the-art. At last, we perform cross-dataset evalution to verify the generalization of the proposed method. 

\subsection{Datasets and Evaluation Protocols}
We conduct extensive experiments on three challenging video-based person re-ID datasets, including the ILIDS-VID~\cite{wang2014person}, PRID 2011~\cite{hirzer2011person} and MARS~\cite{zheng2016mars}.~\textbf{The ILIDS-VID dataset} contains 300 identities forming a total number of 600 video sequences. Pedestrians are observed in two non-overlapping cameras views. The length of each tracklet varies from 23 to 192, with an average number of 73 frames. The train and test set are splited evenly of 150 identities.~\textbf{The PRID 2011 dataset} is another standard benchmark for video-based person re-ID. Following~\cite{dai2019video,liu2018spatial}, we select 178 out of 200 identities with more than 21 frames forming a total number of 356 video sequences.~\textbf{The MARS dataset} is the one of the largest datasets, which consists of 1,261 different IDs and around 20,000 tracklets from 6 cameras. We evaluate the performance of the proposed method on ILIDS-VID and MARS, and implement cross-dataset evaluation on PRID2011. 
% In addition, we also train and test our framework on UCF-101~[cite] which is a widely used action recognition dataset.
    
We evaluate our model by Mean Average Precision (mAP) score and Cumulative Matching Characteristic (CMC) curve.
% For the MARS dataset, we adopt mAP score and the CMC curve at Rank-1, Rank-5, Rank-10, Rank-20 as the evaluation metric. For both PRID 2011 and ILIDS-VID, Rank-1, Rank-5, Rank-10 and Rank-20 score of the CMC curve are reported. We employ the accuracy for UCF-101.\\
 \begin{figure}[t]
\includegraphics[width=0.48\textwidth] {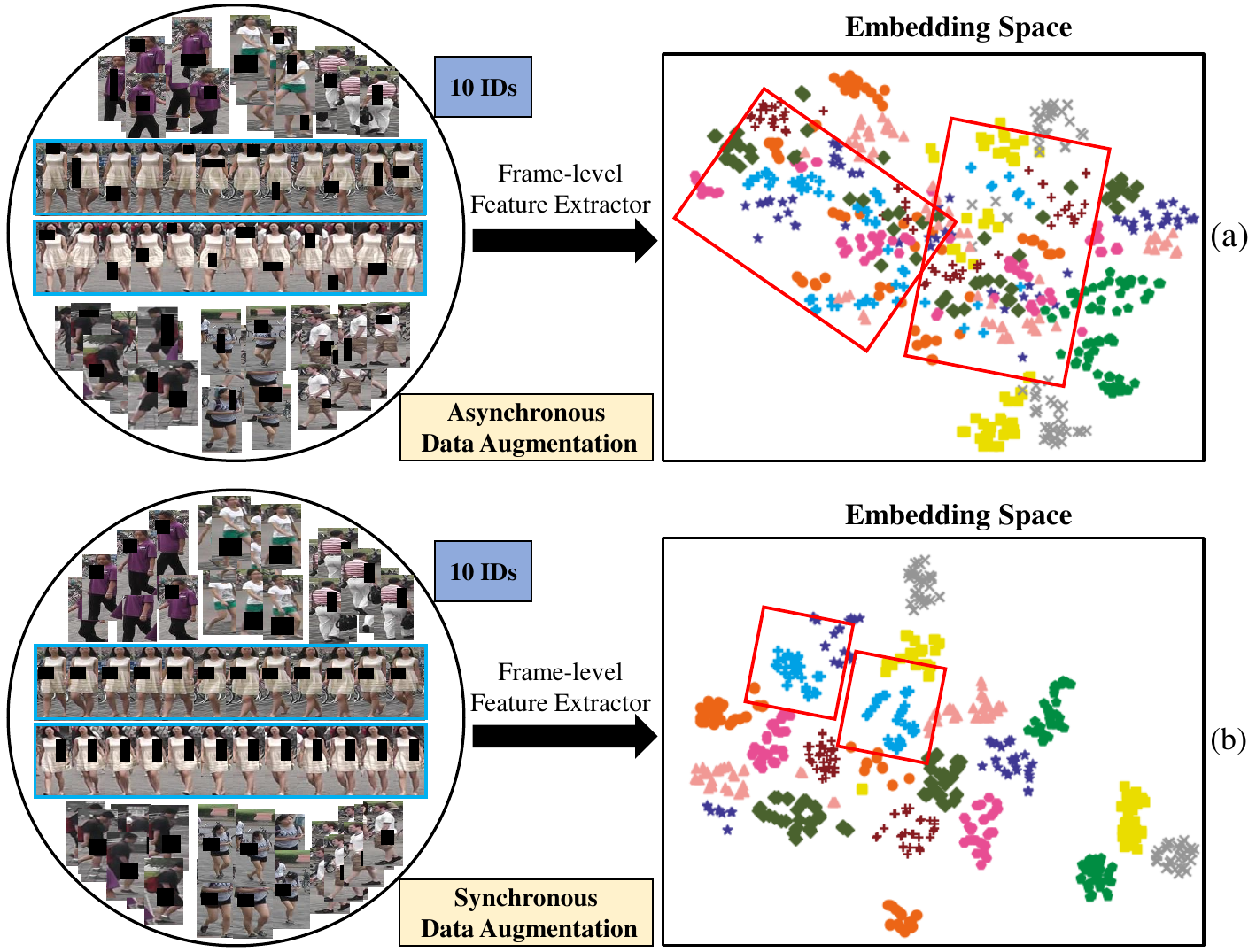}
\caption{Visualization of the embedding space. We sampled ten identities, and each ID contains two video clips with twelve frames. We use ResNet50, pre-trained on ImageNet, as frame-level feature extractor, and then visualize these features by t-SNE~\cite{maaten2008visualizing}.}
\label{tsne}
\end{figure}

\subsection{ Implementation Details}
 Our model is supervised by triplet loss~\cite{hermans2017defense} and cross-entropy loss~\cite{bishop2006pattern}.
%  For ICA module, the weights of non-linear block are initialized with kaiming initialization. 
 Adam~\cite{kingma2014adam} optimizer with $\beta_1=0.9$ and $\beta_2=0.999$ is adopted to optimize the proposed framework, where weight decay is set to $5\times10^{-4}$. The learning rate is initially set to $4\times10^{-4}$ and later down-scaled by a factor of $\frac{1}{200}$ every 200 epochs until 500 epochs. We randomly sample 4 identities with 8 consecutive clips forming the mini-batch of 32. According to~\cite{gao2018revisiting,fu2019sta}, we set the length of each clip to 4. The input frames are uniformly resized to $256\times 128\times3$ and linearly scaled to [-1,1]. In training, the temporally random cropping, temporally random flipping and temporally random erasing are used to augment each video clip. We adopt the same settings for training all datasets. We utilize the ResNet50~\cite{he2016deep} pre-trained on ImageNet as the frame-level feature extraction network. Note that out method can be easily generalized to other backbones. All experiments are conducted on a server with Python 3.6.4 and Pytorch 1.1 platform. 
%  We use the same settings for training all dataset except for UCF-101.

% Each input sequence is augmented with synchronized transformation and re-sized the all frames of a given sequence to 7
% 24$\times$112$\times$3. We use ImagegNet pretrained model for the base network. We set a smaller learning rate in our experiment, which initialized as 0.0004 and decreases to its $\frac{1}{200}$ every 200 epochs. The weight of ICA and classifier are initialized the same as ResNet strategy~\cite{he2015delving}. We use both triplet loss and cross-entropy loss~\cite{bishop2006pattern} to train out network. During testing, we not use synchronized transformation techniques, except for re-size the input frames to 256$\times$128$\times$3. To calculate the similarity between query and gallery video, we average all clip level feature vector $f_p^c$ one IDs to represent the IDs feature vector $\Bar{f_p}$, and then compute the metric distance with query feature vector $\Bar{q}$. To prove the generality of our method, cross-data evaluation is conducted on PRID2011 by utilized the best model trained on ILIDS-VID.
\begin{table}[t]
% \begin{center}
\centering
\scalebox{0.7}{
\begin{tabular}{l|c|lll|ccc|c}
% \specialrule{0em}{0pt}{0pt}
\hline
\multicolumn{1}{l|}{Datasets} & \multirow{2}{*}{Model}  & \multicolumn{3}{c|}{ILIDS-VID} & \multicolumn{4}{c}{MARS} \\ \cline{1-1} \cline{3-9}
 \multicolumn{1}{l|}{Rank@k} & & \multicolumn{1}{c}{1} & \multicolumn{1}{c}{5}  & \multicolumn{1}{c|}{20} & \multicolumn{1}{c}{1} & \multicolumn{1}{c}{5}  & \multicolumn{1}{c|}{20} & \multicolumn{1}{c}{mAP} \\ \hline \hline 
\multicolumn{1}{l|}{Baseline} & Model 1  &78.2  &87.8   &92.1  &77.8  &86.4  &90.7  &74.5 \\
\multicolumn{1}{l|}{Baseline + SDA}  &Model 2  &80.3  &89.5  &92.8  &79.1  &87.4  &91.4  &75.6  \\
\multicolumn{1}{l|}{ICA} &Model 3  &86.1  &96.9   &98.7  &85.1  &95.2  &97.4  &80.7 \\
\multicolumn{1}{l|}{\bf{ICA + SDA}}  &Model 4 &\bf{88.7}  &\bf{98.7}  &\bf{100.0}  &\bf{87.5}  &\bf{96.6}  &\bf{98.2}  &\bf{81.6}  \\
\hline
\end{tabular}}
% \end{center}
\caption{Ablation on the effectiveness of the proposed components.}
\label{componets}
\end{table}

\begin{table}[t]
\centering
\scalebox{0.62}{
\begin{tabular}{c|c|ccc|ccc|c}
\hline
%\toprule[0.4pt]
\multicolumn{1}{l|}{Datasets} & \multirow{2}{*}{Backbone}  & \multicolumn{3}{c|}{ILIDS-VID} & \multicolumn{4}{c}{MARS} \\ \cline{1-1} \cline{3-9}
\multicolumn{1}{l|}{Rank@k} &  & \multicolumn{1}{c}{1} & \multicolumn{1}{c}{5} & \multicolumn{1}{c|}{20} & \multicolumn{1}{c}{1}  & \multicolumn{1}{c}{5} & \multicolumn{1}{c|}{20} & \multicolumn{1}{c}{mAP} \\ \hline \hline
\multirow{3}{*}{Baseline + SDA}    & Alexnet~\cite{krizhevsky2012imagenet}      &53.1  &69.4 &75.1 &51.3 &66.8  &73.8  &49.7 \\
 & InceptionV3~\cite{szegedy2016rethinking}  &69.5  &83.6 &89.7  &67.2  &81.6  &87.3 &62.5   \\ 
 & ResNet50~\cite{he2016deep} &80.3  &89.5  &92.8  &79.1  &87.4  &91.4  &75.6  \\ \hline
\multirow{3}{*}{ICA + SDA}    & Alexnet~\cite{krizhevsky2012imagenet}      &62.1  &79.0 &83.1 &59.5 &76.1  &80.3  &54.7 \\
& InceptionV3~\cite{szegedy2016rethinking}  &78.4  &94.1  &97.4  &76.3  &91.4  &95.0  &72.0    \\ 
& ResNet50~\cite{he2016deep} &\bf{88.7}  &\bf{98.7}  &\bf{100.0}  &\bf{87.5}  &\bf{96.6}  &\bf{98.2}  &\bf{81.6}  \\ 
\hline
%\bottomrule[0.4pt]
 \end{tabular}}
% \end{center}
\caption{Ablation on the effectiveness of different feature extractors.}
\label{extractor}
\end{table}

%-------------------------------------------------------------------------
\subsection{Ablation Study}
% In this part, we conduct ablation investigation to analyze the effect of several factors upon the performance, which include the ICA module, the setting of synchronized transformation.\\
% In this subsection, we comprehensively study the contribution and generality of the proposed components on ILIDS-VID, MARS and PRID 2011 dataset.

The baseline (Model 1) corresponds to resnet-50 with temporal average pooling, and training with asynchronously random cropping, flipping and erasing. The effectiveness of each component is reported in Table~\ref{componets}. From top to bottom, we evaluate each component successively. We can observe that, with the proposed video-based data augmentation, Model 2 surpasses the Model 1 on ILIDS-VID and MARS datasets about 2.1\% and 1.3\% at rank-1, respectively. Attributed to ICA module, Model 3 improves the rank-1 accuracy by 7.9\% and 7.3\% than Model 1 on ILIDS-VID and MARS, respectively. By combining the proposed SDA and ICA module, Model 4 works better than other models by a large margin. We also implement our method on different backbones, and the result shows that “ICA+SDA” consistently outperforms baseline, demonstrating that our method performs well with different frame feature extractors.
% the performance are further boosted than other models with a large margin. 
%  The result indicates that hierarchical aggregation scheme full integrated the intra-clip information to generate more discriminative descriptors of a tracklet.
% , which convincingly demonstrates that the video sequence with sufficient complementary information can exactly improves the representation ability of CNN. 

% To clarify the generality of each component, we conduct cross-data evaluation by using the above models trained on MARS and performing inference on PRID 2011. As shown in Tab.2, Model 2 yields consistent improvement on the new domain, which convincingly demonstrates that preserving more consistently temporal information is benefit to boost the CNN representation ability. In addition, Model 3 outperforms the baseline about 9.7\% and 10.7\% at rank-1 in Tab.1 and Tab.2. The result demonstrates the effectiveness and generality of the ICA model.
% Model 3 shows a performance advantage of 10.6\%, 14.1\% and 9.9\% at rank-1, 5, and 20 over the Model 1.
 To better understand the difference between asynchronous and synchronous data augmentation, we carefully visualize the corresponding embedding space, as shown in Fig.~\ref{tsne}. We can see that the (a) illustrates a scattered distribution, where the intra-clip frames are separated and mixed with other clips. The (b) shows intra-clip frames are clustered and preserve more clip-level information.
\subsection{Comparison with State-of-the-arts}
We use the best model (Model 4) obtained by the proposed ICA and SDA to compare with previous state-of-the-art results. As shown in Table~\ref{compare}, the first two methods~\cite{liu2017quality,liu2018video} explicitly utilize the temporal information by using a on-line network to estimate the optical flow between the consecutive frames and combining them with the spatial information to represent the clip-level features. The third method~\cite{liao2018video} contributes to implicitly use the spatio-temporal information with a succession of 3D convolutions. The last seven methods~\cite{dai2019video,su2018spatial,li2018diversity,gao2018revisiting,liu2018spatial,fu2019sta,li2019global} aggregate intra-clip features over temporal dimension to represent the clip-level features. \textbf{Note:} STMP~\cite{liu2018spatial} uses inceptionV3 as their backbone network. Different feature extractors have significantly impact on the final performance, as shown in Table~\ref{extractor}. We carefully re-implement~\cite{liu2018spatial} by adopting ResNet50 as the backbone network. 
Table~\ref{compare} reveals that our method outperforms other state-of-the-art methods by a large margin, more than 0.5\% better than the recent proposed STA~\cite{fu2019sta} and GLTR~\cite{li2019global} at rank-1 and rank-5 on MARS benchmark.
Due to the over-fitting trap, the previous methods may exhibit good performance on the training set, but suffers severe performance degeneration when work on a new dataset. To better understand the generalization performance of our method, we conducted cross-dataset experiment. We trained Model 4 on ILIDS-VID and evaluate it on PRID2011. Table~\ref{generality} shows that our model achieves consistently superior performance over other methods, which demonstrates the generality of our method.

\begin{table}[t]
\centering
\scalebox{0.78}{
\begin{tabular}{lll|ccc|ccc|c}
\hline
%\toprule[0.4pt]
\multicolumn{3}{l|}{Datasets} & \multicolumn{3}{c|}{ILIDS-VID} & \multicolumn{4}{c}{MARS} \\ \hline
\multicolumn{3}{l|}{Rank@k} & \multicolumn{1}{c}{1} & \multicolumn{1}{c}{5} & \multicolumn{1}{c|}{20} & \multicolumn{1}{c}{1} & \multicolumn{1}{c}{5} & \multicolumn{1}{c|}{20} & \multicolumn{1}{c}{mAP} \\ \hline \hline
\multicolumn{3}{l|}{QAN~\cite{liu2017quality}}           &68.0  &86.8  &97.4  &73.7  &84.9  &91.6  &51.7     \\
\multicolumn{3}{l|}{AMOC~\cite{liu2018video}} &68.7  &94.3  &99.3  &68.3  &81.4  &90.6  &52.9     \\ 
\multicolumn{3}{l|}{Liao~\textit{et al.}~\cite{liao2018video}} &81.3  &{-}   &{-}   &84.3  &{-}   &{-}   &77.0     \\
\multicolumn{3}{l|}{TRL~\cite{dai2019video}}     &57.7  &81.7  &94.1  &79.3  &91.1  &96.0  &66.8  \\
\multicolumn{3}{l|}{STSRN~\cite{su2018spatial}}         &70.0  &89.3  &98.7  &76.7  &93.8  &98.1  &{-}     \\
\multicolumn{3}{l|}{DRSA~\cite{li2018diversity}}&80.2  &{-}   &{-}   &82.3  &{-}   &{-}   &65.8     \\
\multicolumn{3}{l|}{Gao~\textit{et al.}~\cite{gao2018revisiting}}        &{-}   &{-}   &{-}   &83.3  &93.8  &97.4  &76.7     \\
\multicolumn{3}{l|}{STMP~\cite{liu2018spatial}}      &85.7  &97.5  &99.8  &86.2  &95.3  &97.8  &75.6     \\

\multicolumn{3}{l|}{STA~\cite{fu2019sta}}      &{-}  &{-}  &{-}   &86.3   &95.7   &97.1
  &80.8     \\
\multicolumn{3}{l|}{GLTR~\cite{li2019global}}      &86.0  &98.0  &{-}   &87.0  &95.8   &98.2
  &78.5     \\
\hline
\multicolumn{3}{l|}{\bf Ours} &\bf{88.7}  &\bf{98.7}  &\bf{100.0}  &\bf{87.5}  &\bf{96.6}  &\bf{98.2}  &\bf{81.6}  \\
\hline
%\bottomrule[0.4pt]
\end{tabular}}
% \end{center}
\caption{Performance comparison with other stare-of-the-art methods on ILIDS-VID and MARS datasets. ``-'': no reported results.}
\label{compare}
\end{table}
%----------------------------------------------------------------------------

\begin{table}[t]
% \begin{center}
\centering
\scalebox{0.85}{
\begin{tabular}{l|l|l|ccc}
\hline
\multicolumn{3}{l|}{Datasets} & \multicolumn{3}{c}{PRID 2011} \\ \hline
\multicolumn{3}{l|}{Rank@k} &~~~~~~1~~~~~~ &~~~~~~5~~~~~~  &~~~~~~20~~  \\ \hline \hline
% \multicolumn{3}{l|}{CNN-RNN~\cite{mclaughlin2016recurrent}} &~~~~~~28.0~~~~~~  &~~~~~~57.0~~~~~~  &~~~~~~81.0~~  \\
\multicolumn{3}{l|}{TRL~\cite{dai2019video}} &~~~~~~29.5~~~~~~ &~~~~~~59.4~~~~~~   &~~~~~~82.2~~  \\
\multicolumn{3}{l|}{STSRN~\cite{su2018spatial}} &~~~~~~30.0~~~~~~  &~~~~~~58.0~~~~~~   &~~~~~~85.0~~  \\ 
\multicolumn{3}{l|}{STMP~\cite{liu2018spatial}} &~~~~~~32.0~~~~~~  &~~~~~~58.0~~~~~~    &~~~~~~90.0~~  \\\hline 
% \multicolumn{3}{l|}{Baseline} &~~~~~~29.7~~~~~~  &~~~~~~57.6~~~~~~   &~~~~~~82.1~~  \\ \hline
% \multicolumn{3}{l|}{w/ ST} &~~~~~~30.0~~~~~~ &~~~~~~57.9~~~~~~    &~~~~~~82.1~~  \\
% \multicolumn{3}{l|}{w/ ICA} &~~~~~~40.3~~~~~~  &~~~~~~70.4~~~~~~   &~~~~~~92.0~~  \\
\multicolumn{3}{l|}{\bf Ours} &\bf ~~~~~~41.6~~~~~~  &\bf ~~~~~~71.9~~~~~~    &\bf ~~~~~~92.1~~  \\ \hline
\end{tabular}}
% \end{center}
\caption{Generality comparison with other state-of-the-arts methods. }
\label{generality}
\end{table}

\section{Conclusions}
In this paper, we revisit data augmentation for video person re-ID task, and propose a video-based data augmentation scheme, termed as Synchronous Data Augmentation, for training the convolutional neural network. Benefited from the proposed data augmentation strategy, our model is better to utilize the interactive information among frames and has strong generality. In order to extract clip-level semantic features, we also propose a ICA module to integrate the intra-clip features in hierarchical manner. Thanks to the proposed data augmentation strategy and temporal integration pipeline, we achieve new state of the art on ILIDS-VID and MARS benchmarks without re-ranking. We also perform the cross-dataset validation and confirm the generality of our method.
% In order to effectively integrate the multi-frame information,

% In training, the proposed data augmentation process each input frame with same probability and operation. 
% By incorporating the proposed data augmentation and ICA module, our pipeline outperforms most state-of-the-arts on ILIDS-VID and MARS benchmarks without re-ranking. 

% we have presented a novel video person re-identification pipeline, which mainly contains two parts: 1) a video-based data augmentation, 2) cascaded intra-clip aggregation module. 
% Our pipeline better use of the interactive information among frames to represent a discriminative clip-level features. For the  
% In this paper, we focus on the problem of unaligned intra-clip data initialization and distinctive clip-level feature representation for video-based person re-id. To handle the above-mentioned problem,we design a novel framework for video-based person re-id, which consists of two main modules: Synchronized Transformation (ST) and Intra-clip Aggregation (ICA). The former module augments intra-clip frames with the same probability and the same operation, while the latter leverages two-level intra-clip encoding to generate more discriminative clip-level features. We conduct ablation study to analyze the effects of each module. Extensive experiments conducted on two benchmarks including MARS and ILIDS-VID demonstrate the synchronized transformation and ICA module are beneficial to intra-clip aggregation. Furthermore, we perform the cross-dataset evalution with our best model to show the generality of our method. 

\bibliographystyle{IEEEbib}
\bibliography{refs}
\end{document}